\documentclass[runningheads]{llncs}
\usepackage{graphicx}

\usepackage{tikz}
\usepackage{comment}
\usepackage{amsmath,amssymb} 
\usepackage{color}
\usepackage{cite}
\usepackage{booktabs}
\usepackage{tabularx}
\usepackage{fixltx2e}
\usepackage[accsupp]{axessibility}  

\newcommand{\etal}{\textit{et al}.}
\DeclareRobustCommand{\quartiles}[3]{#2\hspace{1pt}\rlap{\textsubscript{#1}}{\textsuperscript{\raisebox{1pt}{#3}}}}

\DeclareMathOperator*{\argmin}{arg\,min}

\begin{document}
\pagestyle{headings}
\mainmatter
\def\ECCVSubNumber{1567}  

\title{CPO: Change Robust Panorama to \\ Point Cloud Localization} 

\titlerunning{CPO: Change Robust Panorama to \\ Point Cloud Localization}
%
\author{Junho Kim\inst{1} \and
Hojun Jang\inst{1} \and
Changwoon Choi\inst{1} \and
Young Min Kim\inst{1,2}}
%
\authorrunning{J. Kim et al.}
%
\institute{Department of Electrical and Computer Engineering, Seoul National University \and
Interdisciplinary Program in Artificial Intelligence and INMC, Seoul National University}
\maketitle

\begin{abstract}
We present CPO, a fast and robust algorithm that localizes a 2D panorama with respect to a 3D point cloud of a scene possibly containing changes.
To robustly handle scene changes, our approach deviates from conventional feature point matching, and focuses on the spatial context provided from panorama images.
Specifically, we propose efficient color histogram generation and subsequent robust localization using score maps.
By utilizing the unique equivariance of spherical projections, we propose very fast color histogram generation for a large number of camera poses without explicitly rendering images for all candidate poses.
We accumulate the regional consistency of the panorama and point cloud as 2D/3D score maps, and use them to weigh the input color values to further increase robustness.
The weighted color distribution quickly finds good initial poses and achieves stable convergence for gradient-based optimization.
CPO is lightweight and achieves effective localization in all tested scenarios, showing stable performance despite scene changes, repetitive structures, or featureless regions, which are typical challenges for visual localization with perspective cameras.
Code is available at \url{https://github.com/82magnolia/panoramic-localization/}.

\keywords{Visual Localization, Panorama, Point Cloud}
\end{abstract}

\section{Introduction}
\label{sec:intro}

\begin{figure}[t]
  \centering
   \includegraphics[width=\linewidth]{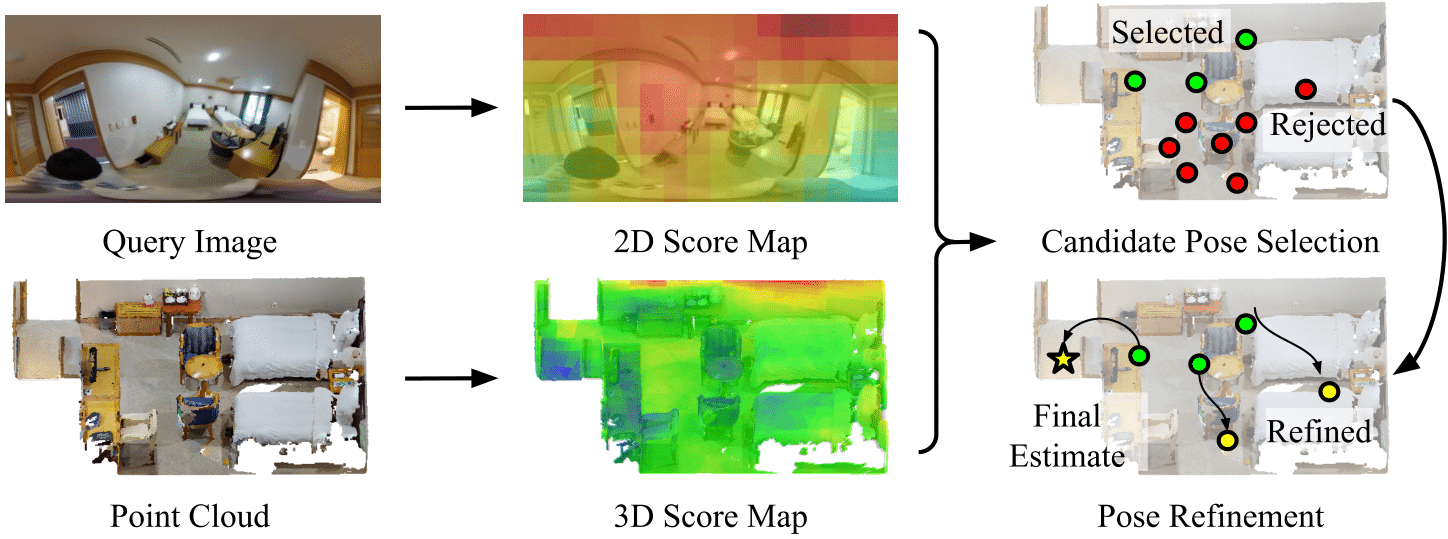}

   \caption{Overview of our approach. CPO first creates 2D and 3D score maps that attenuate regions containing scene changes. The score maps are further used to guide candidate pose selection and pose refinement.}
   \label{fig:overview}
\end{figure}

The location information is a crucial building block to develop applications for AR/VR, autonomous driving, and embodied agents.
Visual localization is one of the cheapest methods for localization as it could operate only using camera inputs and a pre-captured 3D map.
While many existing visual localization algorithms utilize perpsective images~\cite{active_search_eccv,sarlin2019coarse,lstm_vis_loc}, they are vulnerable to repetitive structures, lack of visual features, or scene changes.
Recently, localization using panorama images~\cite{gosma, gopac, piccolo, sphere_cnn} has gained attention, as devices with $360^\circ$ cameras are becoming more accessible.
The holistic view of panorama images has the potential to compensate for few outliers in localization and thus is less susceptible to minor changes or ambiguities compared to perspective images.

Despite the potential of panorama images, it is challenging to perform localization amidst drastic scene changes while simultaneously attaining  efficiency and accuracy.
On the 3D map side, it is costly to collect the up-to-date 3D map that reflects the frequent changes within the scenes.
On the algorithmic side, existing localization methods have bottlenecks either in computational efficiency or accuracy.
While recent panorama-based localization methods~\cite{gosma, gopac, piccolo, sphere_cnn} perform accurate localization by leveraging the holistic context in panoramas, they are vulnerable to scene changes without dedicated treatment to account for changes.
For perspective cameras, such scene changes are often handled by a two-step approach, using learning-based robust image retrieval~\cite{netvlad, openibl} followed by feature matching~\cite{sarlin2020superglue}.
However, the image retrieval step involves global feature extraction which is often costly to compute and memory intensive.

\begin{figure*}[t]
  \centering
   \includegraphics[width=\linewidth]{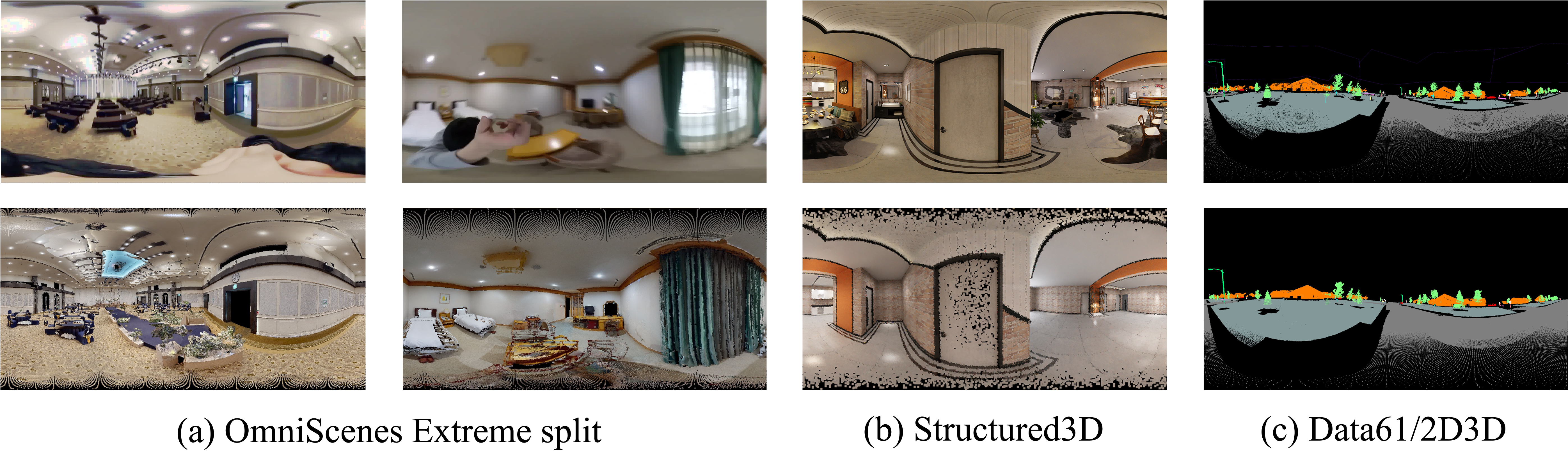}
   \vspace{-2em}
   \caption{Qualitative results of CPO. We show the query image (top), and the projected point cloud on the estimated pose (bottom). CPO can flexibly operate using raw color measurements or semantic labels.}
   \label{fig:qualitative}
   \vspace{-1.5em}
\end{figure*}

We propose CPO, a fast localization algorithm that leverages the regional distributions within the panorama images for robust pose prediction under scene changes.
Given a 2D panorama image as input, we find the camera pose using a 3D point cloud as the reference map.
With careful investigation on the pre-collected 3D map and the holistic view of the panorama, CPO focuses on regions with consistent color distributions.
CPO represents the consistency as 2D/3D score maps and quickly selects a small set of initial candidate poses from which the remaining discrepancy can be quickly and stably optimized for accurate localization as shown in Figure~\ref{fig:overview}.
As a result, CPO enables panorama to point cloud localization under scene changes without the use of pose priors, unlike the previous state-of-the-art~\cite{piccolo}.
Further, the formulation of CPO is flexible and can be applied on both raw color measurements and semantic labels, which is not possible with conventional structure-based localization relying on visual features.
To the best of our knowledge, we are the first to explicitly propose a method for coping with changes in panorama to point cloud localization.

The key to fast and stable localization is the efficient color histogram generation that scores the regional consistency of candidate poses.
Specifically, we utilize color histograms generated from synthetic projections of the point cloud and make comparisons with the query image.
Instead of extensively rendering a large number of synthetic images, we first cache histograms in a few selected views.
Then, color histograms for various other views are efficiently approximated by re-using the histograms of the nearest neighbor from the pre-computed color distribution of overlapping views.
As a result, CPO generates color histograms for millions of synthetic views within a matter of milliseconds and thus can search a wide range of candidate poses within an order-of-magnitude shorter runtime than competing methods.
We compare the color histograms and construct the 2D and 3D score maps, as shown in Figure~\ref{fig:overview} (middle).
The score maps impose higher scores in regions with consistent color distribution, indicating that the region did not change from the reference 3D map.
The 2D and 3D score maps are crucial for change-robust localization, which is further verified with our experiments.

We test our algorithm in a wide range of scenes with various input modalities where a few exemplar results are presented in Figure~\ref{fig:qualitative}.
CPO outperforms existing approaches by a large margin despite a considerable amount of scene change or lack of visual features.
Notably, CPO attains highly accurate localization, flexibly handling both RGB and semantic labels in both indoor and outdoor scenes, without altering the formulation.
Since CPO does not rely on point features, our algorithm  is quickly applicable in an off-the-shelf manner without any training of neural networks or collecting pose-annotated images.
We expect CPO to be a lightweight solution for stable localization in various practical scenarios.

\section{Related Work}
\label{sec:rel}
In this section, we describe prior works for localization under scene changes, and further elaborate on conventional visual localization methods that employ either a single-step or two-step approach.

\paragraph{Localization under Scene Changes}
Even the state-of-the-art techniques for visual localization can fail when the visual appearance of the scene changes.
This is because conventional localization approaches are often designed to find similar visual appearances from pre-collected images with ground-truth poses.
Many visual localization approaches assume that the image features do not significantly change, and either train a neural network~\cite{hierarchical_scene, posenet, lstm_vis_loc, posenet} or retrieve image features~\cite{active_search_eccv, snavely, synth_view_rgbd, synthetic_view_with_matching, active_search}.
Numerous datasets and approaches have been presented in recent years to account for change-robust localization.
The proposed datasets reflect day/night~\cite{lstm_vis_loc,robotcar} or seasonal changes~\cite{cmu_1,cmu_2,robotcar} for outdoor scenes and changes in the spatial arrangement of objects~\cite{inloc,change_vis_2, Structured3D} for indoor scenes.
To cope with such changes, most approaches follow a structure-based paradigm, incorporating a robust image retrieval method~\cite{robust_retrieval, Dusmanu2019CVPR, netvlad, openibl} along with a learned feature matching module~\cite{sarlin2019coarse, sarlin2020superglue, ZhouCVPRpatch2pix, routing}.
An alternative approach utilizes indoor layouts from depth images, which stay constant despite changes in object layouts~\cite{lalaloc}.
We compare CPO against various change-robust localization methods, and demonstrate that CPO outperforms the baselines amidst scene changes.

\paragraph{Single-Step Localization}
Many existing methods~\cite{gosma, gopac, sphere_cnn} for panorama-based localization follow a single-step approach, where the pose is directly found with respect to the 3D map.
Since panorama images capture a larger scene context, fewer ambiguities arise than perspective images, and reasonable localization is possible even without a refinement process or a pose-annotated database.
Campbell \etal~\cite{gosma, gopac} introduced a class of global optimization algorithms that could effectively find pose in diverse indoor and outdoor environments~\cite{stanford2d3d, nicta}.
However, these algorithms require consistent semantic segmentation labels for both the panorama and 3D point cloud, which are often hard to acquire in practice.
Zhang \etal~\cite{sphere_cnn} propose a learning-based localization algorithm using panoramic views, where networks are trained using rendered views from the 3D map.
We compare CPO with optimization-based algorithms~\cite{gopac,gosma}, and demonstrate that CPO outperforms these algorithms under a wide variety of practical scenarios.

\paragraph{Two-Step Localization}
Compared to single-step methods, more accurate localization is often acquired by two-step approaches that initialize poses with an effective search scheme followed by refinement.
For panorama images, PICCOLO~\cite{piccolo} follows a two-step paradigm, where promising poses are found and further refined using sampling loss values that measure the color discrepancy in 2D and 3D.
While PICCOLO does not incorporate learning, it shows competitive performance in conventional panorama localization datasets~\cite{piccolo}.
Nevertheless, the initialization and refinement is unstable to scene changes as the method lacks explicit treatment of such adversaries.
CPO improves upon PICCOLO by leveraging score maps in 2D that attenuate changes for effective initialization and score maps in 3D that guide sampling loss minimization for stable convergence.

For perspective images, many structure-based methods~\cite{sarlin2019coarse, openibl} use a two-step approach, where candidate poses are found with image retrieval~\cite{netvlad} or scene coordinate regression~\cite{hierarchical_scene} and further refined with PnP-RANSAC~\cite{ransac} from feature matching~\cite{sarlin2019coarse, sarlin2020superglue, ZhouCVPRpatch2pix}.
While these methods can effectively localize perspective images, the initialization procedure often requires neural networks that are memory and compute intensive, trained with a dense, pose-annotated database of images.
We compare CPO against prominent two-step localization methods, and demonstrate that CPO attains efficiency and accuracy with an effective formulation in the initialization and refinement.

\section{Method}
\label{sec:method}

Given a point cloud $P=\{X, C\}$,  CPO aims to find the optimal rotation $R^* \in SO(3)$ and translation $t^* \in \mathbb{R}^{3}$ at which the image $I_Q$ is taken.
Let $X, C \in \mathbb{R}^{N\times 3}$ denote the point cloud coordinates and color values, and  $I_Q \in \mathbb{R}^{H\times W \times 3}$ the query panorama image.
Figure~\ref{fig:overview} depicts the steps that CPO localizes the panorama image under scene changes.
First, we extensively measure the color consistency between the panorama and point cloud in various poses.
We propose fast histogram generation described in Section~\ref{sec:fast} for efficient comparison.
The consistency values are recorded as a 2D score map $M_\text{2D} \in \mathbb{R}^{H \times W \times 1}$ and a 3D score map $M_\text{3D} \in \mathbb{R}^{N \times 1}$ which is defined in Section~\ref{sec:score}.
We use the color histograms and score maps to select candidate poses (Section~\ref{sec:candidate}), which are further refined to deduce the final pose (Section~\ref{sec:refine}).

\subsection{Fast Histogram Generation}
\label{sec:fast}

Instead of focusing on point features, CPO relies on the regional color distribution of images to match the global context between the 2D and 3D measurements.
To cope with color distribution shifts from illumination change or camera white balance, we first preprocess the raw color measurements in 2D and 3D via color histogram matching~\cite{hist_match, hist_match_2, C5}.
Specifically, we generate a single color histogram for the query image and point cloud, and establish a matching between the two distributions via optimal transport.
While more sophisticated learning-based methods~\cite{style_transfer_1, style_transfer_2, style_transfer_3} may be used to handle drastic illumination changes such as night-to-day shifts, we find that simple matching can still handle a modest range of color variations prevalent in practical settings.
After preprocessing, we compare the intersections of the RGB color histograms between the \emph{patches} from the query image $I_Q$ and the synthetic projections of the point cloud $P$.

\begin{figure}[t]
    \centering
    \includegraphics[width=0.7\linewidth]{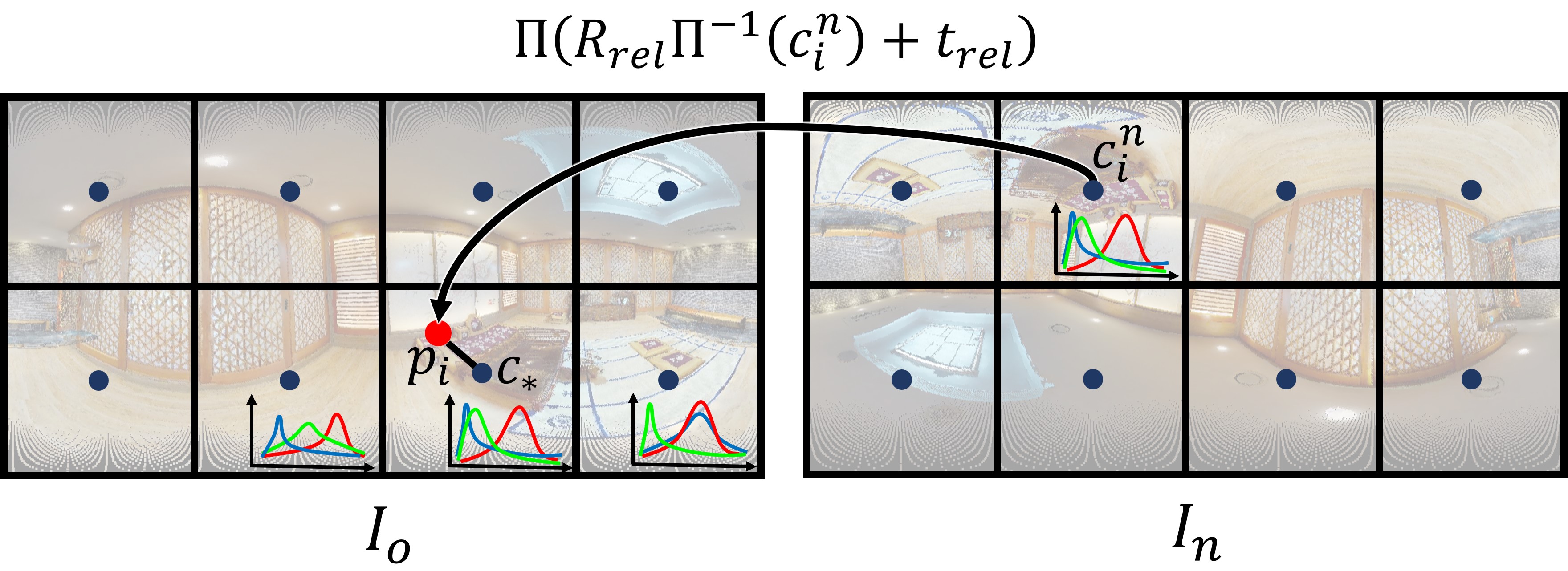}
    \caption{Illustration of fast histogram generation. For each image patch in the novel view $I_n$, we first project the patch centroid $c_i^n$ to the view of the original image $I_o$. The color histogram of the patch in the novel view is estimated as the histogram of image patch $c_*$ in the original view  that is closest to the transformed centroid $p_i$. }
    \label{fig:hist}
\end{figure}

The efficient generation of color histograms is a major building block for CPO.
While there could be an enormous number of poses that the synthetic projections can be generated from, we re-use the pre-computed histograms from another view to accelerate the process.
Suppose we have created color histograms for patches of images taken from the original view $I_o$, as shown in Figure~\ref{fig:hist}.
Then the color histogram for the image in a new view $I_n$ can be quickly approximated without explicitly rendering the image and counting bins of colors for pixels within the patches.
Let $\mathcal{S}_o=\{S^o_i\}$ denote the image patches of $I_o$ and $\mathcal{C}_o=\{c_i^o\}$ the 2D image coordinates of the patch centroids.
$\mathcal{S}_n$ and $\mathcal{C}_n$ are similarly defined for the novel view $I_n$.
For each novel view patch, we project the patch centroid using the relative transformation and obtain the color histogram of the nearest patch of the original image, as described in Figure~\ref{fig:hist}.
To elaborate, we first map the patch centroid location $c_i^n$ of $S_i^n \in \mathcal{S}_n$ to the original image coordinate frame,
\begin{equation}
\label{eq:warp}
    p_i = \Pi(R_\text{rel}\Pi^{-1}(c_i^n) + t_\text{rel}),
\end{equation}
where $R_\text{rel}, t_\text{rel}$ is the relative pose and $\Pi^{-1}(\cdot): \mathbb{R}^{2}\rightarrow \mathbb{R}^{3}$ is the inverse projection function that maps a 2D coordinate to its 3D world coordinate.
The color histogram for $S_i^n$ is assigned as the color histogram of the patch centroid in $I_o$ that is closest to $p_i$, namely $c_* = \argmin_{c \in \mathcal{C}_o} \|c - p_i\|_2$.

We specifically utilize the cached histograms to generate histograms with arbitrary rotations at a fixed translation.
In this case, the camera observes the same set of visible points without changes in occlusion or parallax effect due to depth.
Therefore the synthetic image is rendered only once and the patch-wise histograms can be closely approximated by our fast variant with $p_i = \Pi (R_\text{rel} \Pi^{-1}(c_i^n))$.

\begin{figure}[t]
  \centering
  \includegraphics[width=0.75\linewidth]{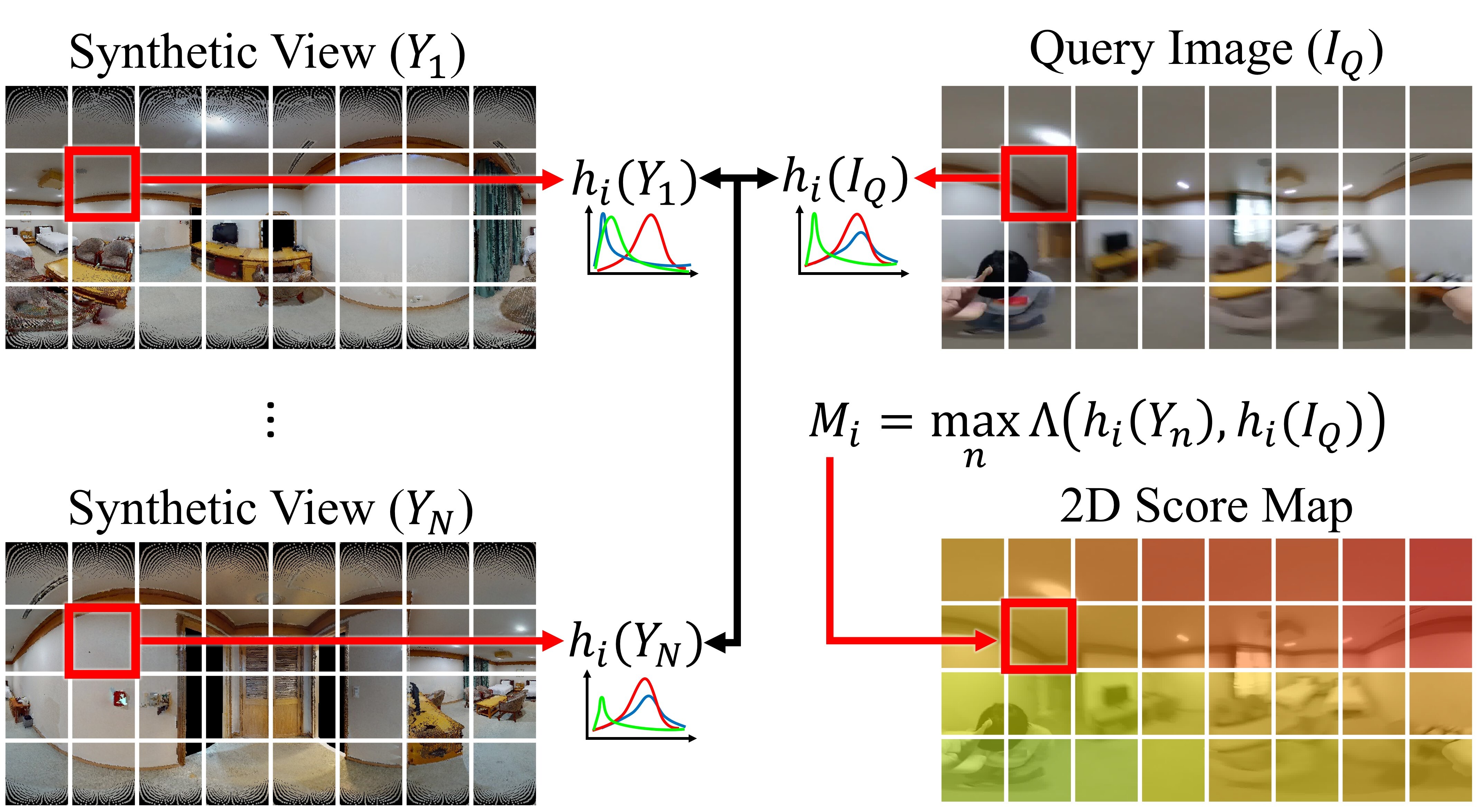}

   \caption{Illustration of 2D score map generation. The 2D score map for the $i$\textsuperscript{th} patch $M_i$ is the maximum histogram intersection between the $i$\textsuperscript{th} patch in query image $I_Q$ and the synthetic views $Y_n \in \mathcal{Y}$.}
   \label{fig:score2d}
   \vspace{-1.5em}
\end{figure}

\subsection{Score Map Generation}
\label{sec:score}

Based on the color histogram of the query image and the synthetic views from the point cloud, we generate 2D and 3D score maps to account for possible changes in the measurements.
Given a query image $I_Q \in \mathbb{R}^{H \times W \times 3}$, we create multiple synthetic views $Y \in \mathcal{Y}$ at various translations and rotations within the point cloud.
Specifically, we project the input point cloud $P=\{X,C\}$ and assign the measured color $Y(u,v)=C_n$  at the projected location of the corresponding 3D coordinate $(u,v)=\Pi(R_Y X_n + t_Y)$ to create the synthetic view $Y$.

We further compare the color distribution of the synthetic views $Y\in \mathcal{Y}$ against the input image $I_Q$ and assign higher scores to regions with high consistency.
We first divide both the query image and the synthetic views into patches and calculate the color histograms of the patches.
Following the notation in Section~\ref{sec:fast}, we can denote the patches of the query image as $\mathcal{S}_Q = \{S_i^Q\}$ and $\mathcal{S}_Y = \{ S_i^Y\}$ for each synthetic view.
Then the color distribution of patch $i$ is recorded into a histogram with $B$ bins per channel: $h_i(\cdot): \mathbb{R}^{H \times W \times 3} \rightarrow S_i \rightarrow \mathbb{R}^{B \times 3}$.
The consistency of two patches is calculated by finding the intersection between two histograms $\Lambda(\cdot, \cdot): \mathbb{R}^{B\times 3} \times \mathbb{R}^{B \times 3} \rightarrow \mathbb{R}$.
Finally, we aggregate the consistency values from multiple synthetic views into the 2D score map for the query image $M_\text{2D}$ and the 3D score map for the point cloud $M_\text{3D}$.
We verify the efficacy of the score maps for CPO in Section~\ref{exp:abl}.

\paragraph{2D Score Map} 
The 2D score map $M_\text{2D} \in \mathbb{R}^{H \times W}$ assigns higher scores to regions in the query image $I_Q$ that are consistent with the point cloud color.
As shown in Figure~\ref{fig:score2d}, we split $M_\text{2D}$ into patches and assign a score for each patch.
We define the 2D score as the maximum histogram intersection that each patch in the input query image $I_Q$ achieves, compared against multiple synthetic views in $\mathcal{Y}$.
Formally, denoting $\mathcal{M}=\{M_i\}$ as the scores for patches in $M_\text{2D}$, the score for the $i^\text{th}$ patch is
\begin{equation}
\label{eq:2d}
    M_i = \max_{Y \in \mathcal{Y}} \Lambda(h_i(Y), h_i(I_Q)).
\end{equation}
If a patch in the query image contains scene change it will have small histogram intersections with any of the synthetic views.
Note that for computing Equation~\ref{eq:2d} we use the fast histogram generation from Section~\ref{sec:fast} to 
avoid the direct rendering of $Y$.
We utilize the 2D score map to attenuate image regions with changes during candidate pose selection in Section~\ref{sec:candidate}.

\begin{figure}[t]
  \centering
    \includegraphics[width=0.9\linewidth]{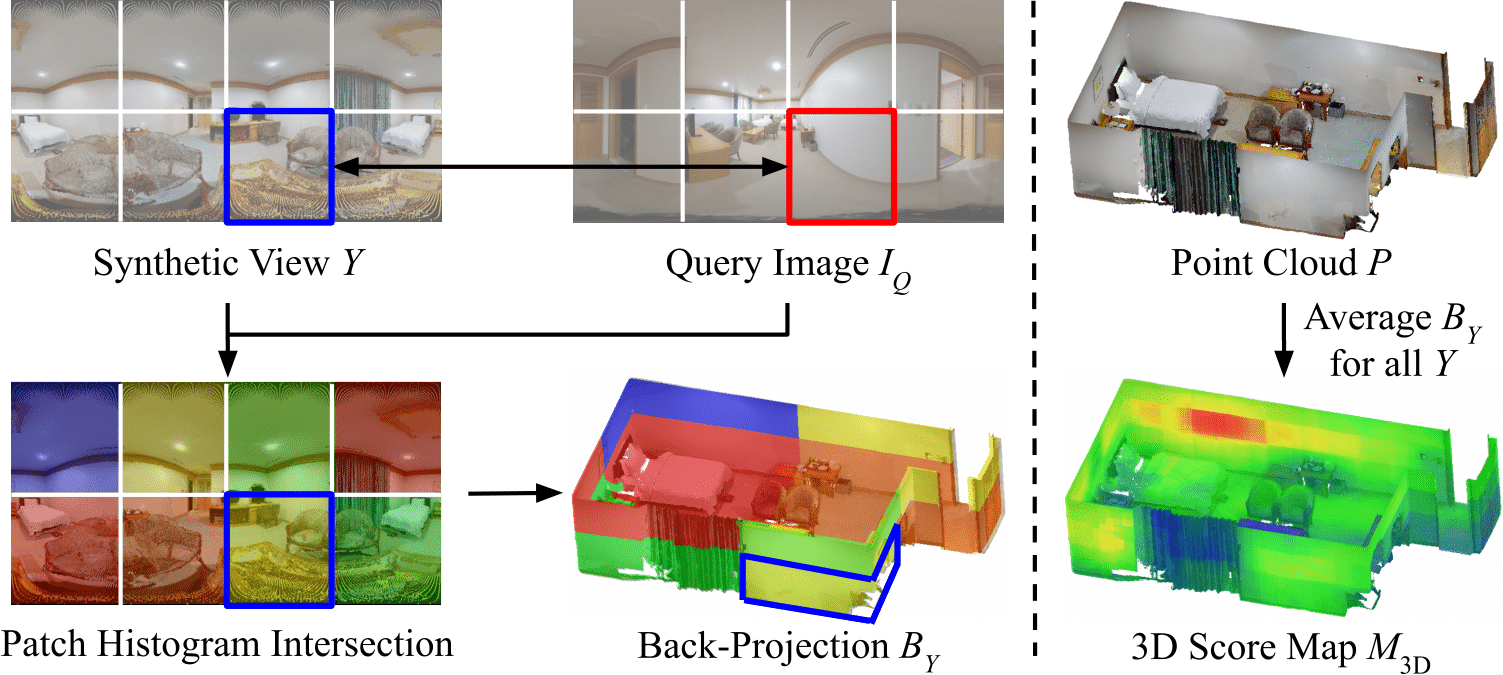}

   \caption{Illustration of 3D score map generation. 
   For each synthetic view $Y \in \mathcal{Y}$, the patch-wise color histogram is compared against the query image and the resulting intersection scores are back-projected onto 3D locations.
   The back-projected scores $B_Y$ are averaged for all synthetic views to form the 3D score map $M_\text{3D}$.}
   \label{fig:score3d}
   \vspace{-1.5em}
\end{figure}

\paragraph{3D Score Map}
The 3D score map $M_\text{3D} \in \mathbb{R}^{N}$ measures the color consistency of each  3D point with respect to the query image.
We compute the 3D score map by back-projecting the histogram intersection scores to the point cloud locations, as shown in Figure~\ref{fig:score3d}.
Given a synthetic view $Y \in \mathcal{Y}$, let $B_Y \in \mathbb{R}^{N}$ denote the assignment of patch-based intersection scores between $Y$ and $I_Q$ into the 3D points whose locations are projected onto corresponding patches in $Y$.
The 3D score map is the average of the back-projected scores $B_Y$ for individual points, namely
\begin{equation}
\label{eq:3d}
    M_\text{3D} = \frac{1}{|\mathcal{Y}|} \sum_{Y \in \mathcal{Y}} B_Y.
\end{equation}
If a region in the point cloud contains scene changes, one can expect the majority of the back-projected scores $B_Y$ to be small  for that region, leading to smaller 3D scores.
We use the 3D score map to weigh the sampling loss for pose refinement in Section~\ref{sec:refine}.
By placing smaller weights on regions that contain scene changes, the 3D score map leads to more stable convergence.

\subsection{Candidate Pose Selection}
\label{sec:candidate}

For the final step, CPO optimizes sampling loss~\cite{piccolo} from selected initial poses, as shown in Figure~\ref{fig:overview}.
CPO chooses the candidate starting poses by efficiently leveraging the  color distribution of the panorama and point cloud.
The space of candidate starting poses is selected in two steps.
First, we choose $N_t$ 3D locations within various regions of the point cloud, and render $N_t$ synthetic views.
For datasets with large open spaces lacking much clutter, the positions are selected from uniform grid partitions.
On the other hand, for cluttered indoor scenes, we propose to efficiently handle valid starting positions by building octrees to approximate the amorphous empty spaces as in Rodenberg \etal~\cite{octree} and select centroids of the octrees for $N_t$ starting positions.

Second, we select the final $K$ candidate poses out of $N_t \times N_r$ poses, where $N_r$ is the number of rotations assigned to each translation, uniformly sampled from $SO(3)$. 
We only render a single view for the $N_t$ locations, and obtain patch-wise histograms for $N_r$ rotations using the fast histogram generation from Section~\ref{sec:fast}.
We select final $K$ poses that have the largest histogram intersections with the query panorama image.
The fast generation of color histograms at synthetic views enables efficient candidate pose selection, which is quantitatively verified in Section~\ref{exp:abl}.

Here, we compute the patch-wise histogram intersections for $N_t \times N_r$ poses where the 2D score map $M_{2D}$ from Section~\ref{sec:score} is used to place smaller weights on image patches that are likely to contain scene change.
Let $\mathcal{Y}_c$ denote the $N_t \times N_r$ synthetic views used for finding candidate poses.
For a synthetic view $Y \in \mathcal{Y}_c$, the weighted histogram intersection $w(Y)$ with the query image $I_Q$ is expressed as follows,
\begin{equation}
\label{eq:intersect}
    w(Y) = \sum_i M_i \Lambda(h_i(Y), h_i(I_Q)).
\end{equation}
Conceptually, the affinity between a synthetic view $Y$ and the query image $I_Q$ is computed as the sum of each patch-wise intersection weighted by the corresponding patch $M_i$ from the 2D score map $M_\text{2D}$.
We can expect changed regions to be attenuated in the candidate pose selection process and therefore CPO can quickly compensate for possible scene changes.

\subsection{Pose Refinement}
\label{sec:refine}
We individually refine the selected $K$ poses by optimizing a weighted variant of sampling loss~\cite{piccolo}, which quantifies the color differences between 2D and 3D.
To elaborate, let $\Pi(\cdot)$ be the projection function that maps a point cloud to coordinates in the 2D panorama image $I_Q$.
Further, let $\Gamma(\cdot;I_Q)$ indicate the sampling function that maps 2D coordinates to pixel values sampled from $I_Q$.
The weighted sampling loss enforces each 3D point's color to be similar to its 2D projection's sampled color while placing lesser weight on points that are likely to contain change.
Given the 3D score map $M_\text{3D}$, this is expressed as follows,
\begin{equation}
    \label{eq:samp_loss}
    L_\mathrm{sampling}(R, t) = \|M_\text{3D} \odot [\Gamma(\Pi(RX+t);I_Q) - C]\|_2,
\end{equation}
where $\odot$ is the Hadamard product and $RX + t$ is the transformed point cloud under the candidate camera pose $R, t$.
To obtain the refined poses, we minimize the weighted sampling loss for the $K$ candidate poses using gradient descent~\cite{adam}.
At termination, the refined pose with the smallest sampling loss value is chosen.

\section{Experiments}
\label{sec:exp}

In this section, we analyze the performance of CPO in various localization scenarios.
CPO is mainly implemented using PyTorch~\cite{pytorch}, and is accelerated with a single RTX 2080 GPU.
We report the full hyperparameter setup for running CPO and further qualitative results for each tested scenario in the supplementary material.
All translation and rotation errors are reported using median values, and for evaluating accuracy a prediction is considered correct if the translation error is below 0.05m and the rotation error is below 5°.

\paragraph{Baselines} We select five baselines for comparison: PICCOLO~\cite{piccolo}, GOSMA~\cite{gosma}, GOPAC~\cite{gopac}, structure-based approach, and depth-based approach.
PICCOLO, GOSMA, and GOPAC are optimization-based approaches that find pose by minimizing a designated objective function.
Structure-based approach~\cite{active_search_eccv, sarlin2019coarse} is one of the most performant methods for localization using perspective images.
This baseline first finds promising candidate poses via image retrieval using global features~\cite{openibl} and further refines pose via learned feature matching~\cite{sarlin2020superglue}.
To adapt structure-based method to our problem setup using panorama images, we construct a database of pose-annotated synthetic views rendered from the point cloud and use it for retrieval.
Depth-based approach first performs learning-based monocular depth estimation on the query panorama image~\cite{pano3d}, and finds the pose that best aligns the estimated depth to the point cloud.
The approach is similar to the layout-matching baseline from Jenkins~\etal~\cite{lalaloc}, where it demonstrated effective localization under scene change.
Additional details about implementing the baselines are deferred to the supplementary material.

\begin{table}[t]
    \centering
    \caption{Quantitative results on all splits containing changes in OmniScenes~\cite{piccolo}.}
    \resizebox{0.9\linewidth}{!}{
    \begin{tabular}{l|ccc|ccc|ccc}
        \toprule
        {} & \multicolumn{3}{c|}{$t$-error (m)} & \multicolumn{3}{c|}{$R$-error ($^\circ$)} & \multicolumn{3}{c}{Accuracy}\\
        Method & Robot & Hand & Extreme & Robot & Hand & Extreme & Robot & Hand & Extreme \\
        \midrule
        PICCOLO & 3.78 & 4.04 & 3.99 & 104.23 & 121.67 & 122.30 & 0.06 & 0.01 & 0.01 \\
        PICCOLO \footnotesize{w/ prior} & 1.07 & 0.53 & 1.24 & 21.03 & 7.54 & 23.71 & 0.39 & 0.45 & 0.38\\
        Structure-Based & 0.04 & 0.05 & 0.06 & \textbf{0.77} & 0.86 & 0.99 & 0.56 & 0.51 & 0.46\\
        Depth-Based & 0.46 & 0.09 &  0.48& 1.35 & 1.24 & 2.37 & 0.38 & 0.39 & 0.30\\
        CPO & \textbf{0.02} & \textbf{0.02} & \textbf{0.03} & 1.46 & \textbf{0.37} & \textbf{0.37} & \textbf{0.58} & \textbf{0.58} & \textbf{0.57}\\
        \bottomrule
    \end{tabular}
    }
    \label{table:omniscenes_change}
\end{table}

\begin{table}[t]
    \centering
    \caption{Quantitative results on all splits containing changes in Structured3D~\cite{Structured3D}.}
    \resizebox{0.9\linewidth}{!}{
    \begin{tabular}{l|cc|ccc}
        \toprule
        Method & $t$-error (m) & $R$-error ($^\circ$) & Acc. (0.05m, $5^\circ$) & Acc. (0.02m, $2^\circ$) & Acc. (0.01m, $1^\circ$) \\
        \midrule
        PICCOLO & 0.19 & 4.20 & 0.47 & 0.45 & 0.43 \\
        Structure-Based &0.02 & 0.64& \textbf{0.59}& 0.47& 0.29\\
        Depth-Based & 0.18 & 1.98 & 0.45 & 0.33 & 0.19 \\
        CPO & \textbf{0.01} & \textbf{0.29} & 0.56 & \textbf{0.54} & \textbf{0.51} \\
        \bottomrule
    \end{tabular}
    }
    \label{table:structured3d}
    \vspace{-1.5em}
\end{table}

\subsection{Localization Performance on Scenes with Changes}
\label{exp:change}
We assess the robustness of CPO using the OmniScenes~\cite{piccolo} and Structured3D ~\cite{Structured3D} dataset, which allows performance evaluation for the localization of panorama images against point clouds in changed scenes.

\paragraph{OmniScenes} The OmniScenes dataset consists of seven 3D scans and 4121 2D panorama images, where the panorama images are captured with cameras either handheld or robot mounted.
Further, the panorama images are obtained at different times of day and include changes in scene configuration and lighting.
OmniScenes contains three splits (Robot, Handheld, Extreme) that are recorded in scenes with changes, where the Extreme split contains panorama images captured with extreme camera motion.

We compare CPO against PICCOLO~\cite{piccolo}, structure-based approach, and depth-based approach.
The evaluation results for all three splits in OmniScenes are shown in Table~\ref{table:omniscenes_change}.
In all splits, CPO outperforms the baselines without the help of prior information or training neural networks.
While PICCOLO~\cite{piccolo} performs competitively with gravity direction prior, the performance largely degrades without such information.
Further, outliers triggered from scene changes and motion blur make accurate localization difficult using structure-based or depth-based methods.
CPO is immune to such adversaries as it explicitly models scene changes and regional inconsistencies with 2D, 3D score maps.

The score maps of CPO effectively attenuate scene changes, providing useful evidence for robust localization.
Figure~\ref{fig:visscore} visualizes the exemplar 2D and 3D score maps generated in the wedding hall scene from OmniScenes.
The scene contains drastic changes in object layout, where the carpets are removed and the arrangement of chairs has largely changed since the 3D scan.
As shown in Figure~\ref{fig:visscore}, the 2D score map assigns smaller scores to new objects and the capturer’s hand, which are not present in the 3D scan.
Further, the 3D score map shown in Figure~\ref{fig:visscore} assigns smaller scores to chairs and blue carpets, which are present in the 3D scan but are largely modified in the panorama image.

\begin{figure}[t]
\centering
\includegraphics[width=\linewidth]{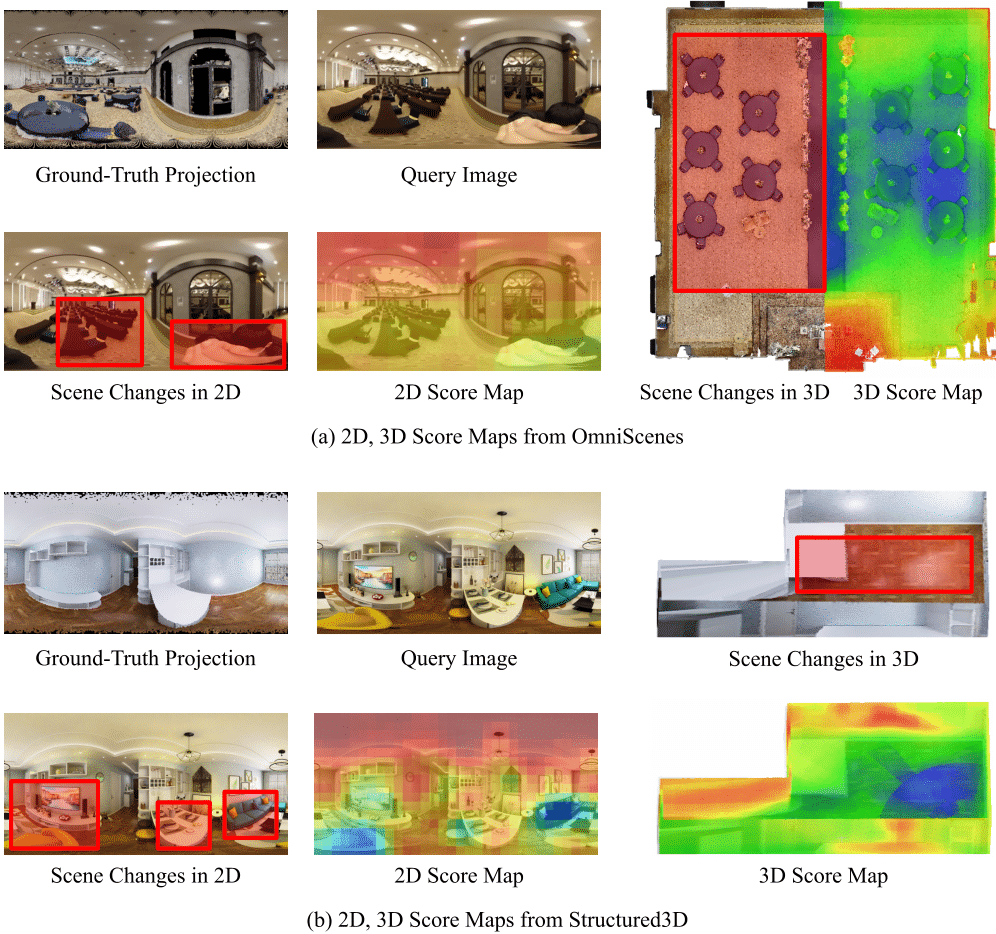}

\caption{Visualization of 2D, 3D score maps in OmniScenes~\cite{piccolo} and Structured3D~\cite{Structured3D}. The 2D score map assigns lower scores to the capturer's hand and  objects not present in 3D. Similarly, the 3D score map assigns lower scores to regions not present in 2D.}
\label{fig:visscore}
\vspace{-1.5em}
\end{figure}

\paragraph{Structured3D} We further compare CPO against PICCOLO in Structured3D, which is a large-scale dataset containing synthetic 3D models with changes in object layout and illumination, as shown in Figure 2.
Due to the large size of the dataset (21845 indoor rooms), 672 rooms are selected for evaluation.
For each room, the dataset contains three object configurations (empty, simple, full) along with three lighting configurations (raw, cold, warm), leading to nine configurations in total.
We consider the object layout change from empty to full, where illumination change is randomly selected for each room.
We provide further details about the evaluation in the supplementary material.
The median errors and localization accuracy at various thresholds is reported in Table~\ref{table:structured3d}.
CPO outperforms the baselines in most metrics, due to the change compensation of 2D/3D score maps as shown in Figure~\ref{fig:visscore}.

\begin{table}[t]
    \centering
    \caption{Quantitative results on Stanford 2D-3D-S~\cite{stanford2d3d}, compared against PICCOLO (PC), structure-based approach (SB), and depth-based approach (DB).}
    \resizebox{\linewidth}{!}{
    \setlength{\tabcolsep}{0.5em}

    \begin{tabular}{l|cccc|cccc|cccc}
        \toprule
        {} & \multicolumn{4}{c|}{$t$-error (m)} & \multicolumn{4}{c|}{$R$-error ($^\circ$)} & \multicolumn{4}{c}{Accuracy}\\
        Area & PC & SB & DB & CPO & PC & SB & DB & CPO & PC & SB & DB & CPO \\
        \midrule
        Area 1 & 0.02 & 0.05 & 1.39 & \textbf{0.01} & 0.46 &  0.81& 89.48 & \textbf{0.25} & 0.66 & 0.51 &0.28 & \textbf{0.89} \\
        Area 2 & 0.76 &  0.18& 3.00 & \textbf{0.01} & 2.25 & 2.08 & 89.76 & \textbf{0.27} & 0.42 & 0.41 & 0.14 & \textbf{0.81} \\
        Area 3 & 0.02 & 0.05 & 1.39 & \textbf{0.01} & 0.49 & 1.01 & 88.94& \textbf{0.24} & 0.53 & 0.50 &0.24 & \textbf{0.76} \\
        Area 4 & 0.18 &  0.05&  1.30& \textbf{0.01} & 4.17 &1.07  & 89.12 & \textbf{0.28} & 0.48 & 0.50 & 0.28 & \textbf{0.83} \\
        Area 5 & 0.50 & 0.10 &2.37 & \textbf{0.01} & 14.64 & 1.31 & 89.88 & \textbf{0.27} & 0.44 & 0.47 & 0.18 & \textbf{0.73} \\
        Area 6 & \textbf{0.01} & 0.04& 1.54 & \textbf{0.01} & 0.31 & 0.74 & 89.39 & \textbf{0.18} & 0.68 & 0.55 & 0.29 & \textbf{0.90} \\
        
        \midrule
        Total & 0.03 & 0.06 & 1.72 & \textbf{0.01} & 0.63 & 1.04& 89.51& \textbf{0.24} & 0.53 &0.49  & 0.23 & \textbf{0.83} \\

        \bottomrule
    \end{tabular}
    }
    \label{table:stanford}
    \vspace{-1.5em}
\end{table}

\subsection{Localization Performance on Scenes without Changes}
\label{exp:prev}

We further demonstrate the wide applicability of CPO by comparing CPO with existing approaches in various scene types and input modalities (raw color / semantic labels).
The evaluation is performed in one indoor dataset (Stanford 2D-3D-S~\cite{stanford2d3d}), and one outdoor dataset (Data61/2D3D~\cite{nicta}).
Unlike OmniScenes and Structured3D, most of these datasets lack scene change.
Although CPO mainly targets scenes with changes, it shows state-of-the-art results in these datasets.
This is due to the fast histogram generation that allows for effective search from the large pool of candidate poses, which is an essential component of panorama to point cloud localization given the highly non-convex nature of the objective function presented in Section 3.

\paragraph{Localization with Raw Color}

We first make comparisons with PICCOLO~\cite{piccolo}, structure-based approach, and depth-based approach in the Stanford 2D-3D-S dataset.
In Table~\ref{table:stanford}, we report the localization accuracy and median error, where CPO outperforms other baselines by a large margin.
Note that PICCOLO is the current state-of-the-art algorithm for the Stanford 2D-3D-S dataset.
The median translation and rotation error of PICCOLO~\cite{piccolo} deviates largely in areas 2, 4, and 5, which contain a large number of scenes such as hallways that exhibit repetitive structure.
On the other hand, the error metrics and accuracy of CPO are much more consistent in all areas.

\begin{table}[t]
    \centering
    \caption{Localization performance using semantic labels on a subset of Area 3 from Stanford 2D-3D-S~\cite{stanford2d3d}. $Q_1$, $Q_2$, $Q_3$ are quartile values of each metric.}
    \resizebox{0.75\linewidth}{!}{
    \setlength{\tabcolsep}{0.5em}
    \begin{tabular}{l|ccc|ccc|ccc}
        \toprule
        {} & \multicolumn{3}{c|}{$t$-error (m)} & \multicolumn{3}{c|}{$R$-error ($^\circ$)} & \multicolumn{3}{c}{Runtime (s)}\\
        {} & $Q_1$ & $Q_2$ & $Q_3$ & $Q_1$ & $Q_2$ & $Q_3$ & $Q_1$ & $Q_2$ & $Q_3$ \\
        \midrule
        PICCOLO & \textbf{0.00} & \textbf{0.01} & 0.07 & \textbf{0.11} & \textbf{0.21} & 0.56 & 14.0 & 14.3 & 16.1 \\
        GOSMA & 0.05 & 0.08 & 0.15 & 0.91 & 1.13 & 2.18 & \textbf{1.4} & 1.8 & 4.4  \\
        CPO & 0.01 & \textbf{0.01} & \textbf{0.02} & 0.20 & 0.32 & \textbf{0.51} & 1.5 & \textbf{1.6} & \textbf{1.6}  \\

        \bottomrule
    \end{tabular}
    }
    \label{table:stanford_area_3}
\end{table}

\begin{table}[t]
    \centering
    \caption{Localization performance on all areas of the Data61/2D3D dataset~\cite{nicta}.}
    \vspace{-0.5em}
    \resizebox{0.6\linewidth}{!}{
    \begin{tabular}{l|ccc|ccc}
        \toprule
        {} & \multicolumn{3}{c|}{$t$-error (m)} & \multicolumn{3}{c}{$R$-error ($^\circ$)}\\
        Method & GOPAC & PICCOLO & CPO & GOPAC & PICCOLO & CPO \\
        \midrule
        Error & 1.1 & 4.9 & \textbf{0.1} & 1.4 & 28.8 & \textbf{0.3} \\
        \bottomrule
    \end{tabular}
    }
    \label{table:nicta}
    \vspace{-1.5em}
\end{table}

\paragraph{Localization with Semantic Labels}

We evaluate the performance of CPO against algorithms that use semantic labels as input, namely GOSMA~\cite{gosma} and GOPAC~\cite{gopac}.
We additionally report results from PICCOLO~\cite{piccolo}, as it could also function with semantic labels.
To accommodate for the different input modality, CPO and PICCOLO use color-coded semantic labels as input, as shown in Figure~\ref{fig:qualitative}(c).
We first compare CPO with PICCOLO and GOSMA on 33 images in Area 3 of the Stanford 2D-3D-S dataset following the evaluation procedure of Campbell \etal~\cite{gosma}.
As shown in Table~\ref{table:stanford_area_3}, CPO outperforms GOSMA~\cite{gosma} by a large margin, with the 3rd quartile values of the errors being smaller than the 1st quartile values of GOSMA~\cite{gosma}.
Further, while the performance gap with PICCOLO~\cite{piccolo} is smaller than GOSMA, CPO consistently exhibits a much smaller runtime.

We further compare CPO with PICCOLO and GOPAC~\cite{gopac} in the Data61/2D 3D  dataset~\cite{nicta}, which is an outdoor dataset that contains semantic labels for both 2D and 3D.
The dataset is mainly recorded in the rural regions of Australia, where large portions of the scene are highly repetitive and lack features as shown in Figure~\ref{fig:qualitative}(c).
Nevertheless, CPO exceeds GOPAC~\cite{gopac} in localization accuracy, as shown in Table~\ref{table:nicta}.
Note that CPO only uses a single GPU for acceleration whereas GOPAC employs a quad-GPU configuration for effective performance~\cite{gopac}.
Due to the fast histogram generation from Section~\ref{sec:fast}, CPO can efficiently localize using a smaller number of computational resources.

\subsection{Ablation Study}
\label{exp:abl}

In this section, we ablate key components of CPO, namely  histogram-based candidate pose selection and 2D, 3D score maps.
The ablation study for other constituents of CPO is provided in the supplementary material.

\paragraph{Histogram-Based Candidate Pose Selection}
We verify the effect of using color histograms for candidate pose selection on the Extreme split from the OmniScenes dataset~\cite{piccolo}.
CPO is compared with a variant that performs candidate pose selection using sampling loss values as in PICCOLO~\cite{piccolo}, where all other conditions remain the same.
As shown in Table~\ref{table:ablation_omniscenes_extreme}, a drastic performance gap is present.
CPO uses patch-based color histograms for pose selection and thus considers larger spatial context compared to pixel-wise sampling loss.
This allows for CPO to effectively overcome ambiguities that arise from repetitive scene structures and scene changes that are present in the Extreme split.

We further validate the efficiency of histogram-based initialization against various initialization methods used in the baselines.
In Table~\ref{table:time}, we report the average runtime for processing a single synthetic view in milliseconds.
The histogram based initialization used in CPO exhibits an order-of-magnitude shorter runtime than other competing methods.
The effective utilization of spherical equivariance in fast histogram generation allows for efficient search within a wide range of poses and quickly generate 2D/3D score maps.

\paragraph{Score Maps}
We validate the effectiveness of the score maps for robust localization under scene changes on the Extreme split from the OmniScenes dataset~\cite{piccolo}.
Recall that we use the 2D score map for guiding candidate pose selection and the 3D score map for guiding pose refinement.
We report evaluation results for variants of CPO that do not use either the 2D or 3D score map.
As shown in Table~\ref{table:ablation_omniscenes_extreme}, optimal performance is obtained by using both score maps.
The score maps effectively attenuate scene changes, leading to stable pose estimation.
\begin{table}[t]
    \centering
    \caption{Ablation of various components of CPO in  OmniScenes~\cite{piccolo} Extreme split.}
    \resizebox{0.6\linewidth}{!}{
    \begin{tabular}{l|ccc}
        \toprule
        Method & $t$-error (m) & $R$-error ($^\circ$) & Acc.\\
        \midrule
        w/o Histogram Initialization & 3.29 & 75.60 & 0.20 \\
        w/o 2D Score Map & 0.10 & 1.19 & 0.48 \\
        w/o 3D Score Map & \textbf{0.03} & 1.56 & 0.55 \\
        \midrule
        Ours & \textbf{0.03} & \textbf{0.37} & \textbf{0.57}\\
        \bottomrule
    \end{tabular}
    }
    \label{table:ablation_omniscenes_extreme}
\end{table}

\begin{table}[t]
    \centering
    \caption{Average runtime for a single synthetic view in Room 3 from OmniScenes~\cite{piccolo}.}
    \resizebox{0.65\linewidth}{!}{
    \begin{tabular}{l|cccc}
        \toprule
        Method & PICCOLO & Structure-Based & Depth-Based & CPO \\
        \midrule
        Runtime (ms) & 2.135 & 38.70 & 2.745 & \textbf{0.188} \\
        \bottomrule
    \end{tabular}
    }
    \label{table:time}
    \vspace{-1.5em}
\end{table}

\section{Conclusion}
\label{sec:conclusion}
In this paper, we present CPO, a fast and robust algorithm for 2D panorama to 3D point cloud localization.
To fully leverage the potential of panoramic images for localization, we account for possible scene changes by saving the color distribution consistency in 2D, 3D score maps.
The score maps effectively attenuate regions that contain changes and thus lead to more stable camera pose estimation.
With the proposed fast histogram generation, the score maps are efficiently constructed and CPO can subsequently select promising initial poses for stable optimization.
By effectively utilizing the holistic context in 2D and 3D, CPO achieves stable localization results across various datasets including scenes with changes.
We expect CPO to be widely applied in practical localization scenarios where scene change is inevitable.

\paragraph{Acknowledgements}
This work was partly supported by the National Research Foundation of Korea (NRF) grant funded by the Korea government(MSIT) (No. 2020R1C1C1008195), Creative-Pioneering Researchers Program through Seoul National University, and Institute of Information \& communications Technology Planning \& Evaluation (IITP) grant funded by the Korea government(MSIT) (No.2021-0-02068, Artificial Intelligence Innovation Hub).


\appendix
\renewcommand\thetable{\thesection.\arabic{table}}    
\setcounter{table}{0}
\renewcommand\thefigure{\thesection.\arabic{figure}}    
\setcounter{figure}{0}

\DeclareRobustCommand{\quartiles}[3]{#2\hspace{1pt}\rlap{\textsubscript{#1}}{\textsuperscript{\raisebox{1pt}{#3}}}}

\pagestyle{headings}
\mainmatter
\def\ECCVSubNumber{1567}  

\title{{\large Supplementary Material for} \\CPO: Change Robust Panorama to Point Cloud Localization} 

\titlerunning{CPO: Change Robust Panorama to \\ Point Cloud Localization}
%
\author{Junho Kim\inst{1} \and
Hojun Jang\inst{1} \and
Changwoon Choi\inst{1} \and
Young Min Kim\inst{1,2}}
%
\authorrunning{J. Kim et al.}
%
\institute{Department of Electrical and Computer Engineering, Seoul National University \and
Interdisciplinary Program in Artificial Intelligence and INMC, Seoul National University}
\maketitle

\section{Structured3D Dataset Details}

\begin{figure}[t]
  \centering
   \includegraphics[width=\linewidth]{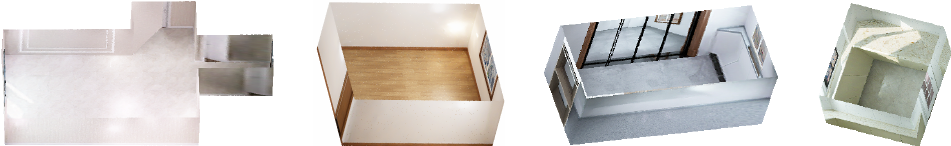}

   \caption{Visualization of synthesized 3D point clouds in  Structured3D~\cite{Structured3D}.}
   \label{fig:model}
\end{figure}

In this section, we provide the details for preparing the Structured3D~\cite{Structured3D} dataset.
Due to copyright constraints, the 3D models of the dataset is unavailable to the public.
Therefore we generated synthetic 3D meshes using the layout annotations and color values from the panorama images, where qualitative samples are shown in Figure~\ref{fig:model}.
As explained in Section 4.1, Structured3D contains 21845 rooms from which we select 672 rooms for evaluation, and each room has three object configurations (empty, simple, full).
We create the 3D model using the empty object layout and set the query panorama as the full object layout.
To additionally evaluate illumination robustness, we randomly choose the lighting setup from the  three possible configurations (raw, cold, warm) for each object configuration.

\section{Baseline Details}
In this section, we describe the details for implementing the baselines compared against CPO.
As we implement PICCOLO~\cite{piccolo} using the publically available codebase released by the authors, we focus our description on the Structure-based and depth-based approaches.
For fair comparison, we set the translation/rotation starting points $N_t, N_r$ and the number of candidate poses $K$ identical to CPO.

\paragraph{Structure-Based Approach}
As explained in Section 4, structured-based approach first finds promising candidate poses using  robust image retrieval and then refines poses using PnP-RANSAC from feature matches.
For image retrieval we use OpenIBL~\cite{openibl}, which is a widely used image retrieval method that outputs a global feature vector for each image.
To deploy OpenIBL in out setup, we first render $N_t \times N_r$ synthetic views from the point cloud.
Then, we extract the global features for each synthetic view and the query image, and choose the top $K$ synthetic views whose feature vectors are closest to that of the query image.
As the final step, we perform feature matching~\cite{sarlin2020superglue} from each chosen synthetic view against the query image, and determine the final view with the most matches.
The pose from the final view is refined with feature matches from the previous step via PnP-RANSAC~\cite{ransac}.

\paragraph{Depth-Based Approach}
Inspired from Jenkins et al.~\cite{lalaloc}, depth-based approach first finds candidate poses by comparing estimated monocular depth with the 3D point cloud and refining pose with PnP-RANSAC.
For monocular depth estimation we use the pretrained model from Albanis et al.~\cite{pano3d}, which can reliably estimate the underlying 3D structure from the query panorama.
Then, we find the top $K$ poses from a pool of $N_t \times N_r$ starting points that have the smallest Chamfer distance with the 3D point cloud.
Similar to the structure-based approach, we perform feature matching and refine the view with the most matches via PnP-RANSAC.

\section{Additional Details on Score Maps}
\label{sec:score}
\subsection{Score Map Generation}

We provide additional details about score map generation.
Recall that we generate 2D, 3D score maps using color consistency from histograms of synthetic views $\mathcal{Y}$.
Here we generate $N^\text{score}_t \times N^\text{score}_r$ synthetic views, similar to the candidate pose selection introduced in Section 3.3.
The exact number of synthetic views used to generate score maps is further specified in Section~\ref{sec:hyper}.

\subsection{2D Score Maps for Pose Refinement}
While not mentioned in the main paper, we empirically found that using both 2D and 3D score maps are helpful during refinement.
This could be explained by the fact that 2D score maps detect outliers in the query image, while 3D score maps detect outliers in the point cloud.
Given a 2D score map $M_{2D}$, we first obtain score values at 2D coordinates from point cloud projections, namely $S_{2D}=\Gamma(\Pi(RX + t); M_{2D})$.
Then, the weighted sampling loss is given as follows,
\begin{equation}
    \label{eq:samp_loss}
    L_\mathrm{sampling}(R, t) = \|(M_\text{3D} + S_{2D}) / 2 \odot [\Gamma(\Pi(RX+t);I_Q) - C]\|_2,
\end{equation}
which is minimized using gradient descent as explained in Section~\ref{sec:refine}.

\section{Hyperparameter Setup}
\label{sec:hyper}

In this section, we report the hyperparameter setups of CPO.
As explained in Section 3.3, from $N_t \times N_r$ poses we select the top $K$ candidate poses with the highest histogram intersection (Equation 4) for pose refinement.
We follow the identical hyperparameter setup as PICCOLO~\cite{piccolo} for pose refinement.
Below we specify other hyperparameter setups that differ by the localization scenario.

\subsection{Localization with Raw Color}
For OmniScenes~\cite{piccolo}, Stanford 2D-3D-S~\cite{stanford2d3d} and Structured3D~\cite{Structured3D}, where localization was done with raw color inputs, we set $N_t=100, K=6$.
We set the number of rotation starting points as $N_r=216$ for OmniScenes and Stanford 2D-3D-S, whereas for Structured3D we use $N_r=24$ to run the baselines in a reasonable amount of time.
For pose selection we split the input image into $8 \times 16$ patches and generate color histograms for each patch using the fast histogram generation presented in Section 3.2.
Other hyperparameter setups slightly differ by dataset, which we elaborate below.

\paragraph{OmniScenes Dataset}
As OmniScenes is mainly an indoor dataset, we employ octree-based translation starting point selection.
For generating 2D and 3D score maps, we use synthetic views from $N^\text{score}_t=100, N^\text{score}_r=216$ poses and divide the input image into $16 \times 32$ patches.
We use patches of finer scale and generate more accurate score maps to cope with large scene changes in OmniScenes~\cite{piccolo}.

\paragraph{Stanford 2D-3D-S}
Similar to OmniScenes, we employ octree-based translation starting point selection, as Stanford 2D-3D-S dataset is also an indoor dataset.
For generating 2D and 3D score maps, we use synthetic views from $N^\text{score}_t=100$, $ N^\text{score}_r=216$ poses and divide the input image into $8 \times 16$ patches.

\paragraph{Structured3D}
As explained in Section A, the 3D models in the Structured3D dataset are synthetically generated cuboids lacking clutter.
Therefore we use a uniform grid partition for this dataset.
Similar to the Stanford 2D-3D-S dataset, for score map generation we use synthetic views from $N^\text{score}_t=100, N^\text{score}_r=24$ poses and divide the input image into $8 \times 16$ patches.

\subsection{Localization with Semantic Labels}
For Stanford 2D-3D-S~\cite{stanford2d3d} and Data61/2D3D~\cite{nicta}, where localization was done with semantic labels, we set $N_r=216$, similar to localization with raw color.
The number of translation starting points $N_t$ differ by dataset, which is further specified below.
In addition, we do not apply score maps in these scenarios as there are no scene changes in both datasets and the color values of semantic labels do not reflect any photometric information.

\paragraph{Stanford 2D-3D-S}
We employ octree-based translation starting point selection and set the number of translation starting points to $N_t=100$, as in raw color localization.
Further, we divide the input image into $8 \times 16$ patches for histogram-based initialization.

\paragraph{Data61/2D3D}
We employ grid-based translation starting point selection and set the number of translation starting points to $N_t=300$, as the dataset is captured outdoor.
Further, we confine the translation domain to a cuboid spanning $50 \times 10 \times 5 m$, similar to the initialization procedure used in Campbell \etal~\cite{gopac}.
The cuboid is placed to cover two lanes within the outdoor scene, which reflects the prior knowledge that the camera was mounted on a vehicle.
For histogram-based initialization, we divide the input image into $4 \times 8$ patches.

\section{Distortion Handling in Histogram Intersection}
\label{sec:distortion}
In this section we describe the distortion handling operation used for calculating histogram intersections in Equation 4.
Since panorama images have spherical distortion, we compensate for such irregularities by applying additional weights proportional to the sin value of the latitude.
To elaborate, we add an additional weight to the histogram intersection equation,
\begin{equation}
\label{eq:intersect}
    w(Y) = \frac{1}{2}\sum_i (M_i+S_i) \Lambda(h_i(Y), h_i(I_Q)),
\end{equation}
where $S_i$ is the sine value of the $i$\textsuperscript{th} patch centroid's latitude.
The modified intersection equation can correctly place lesser weight on patches near the pole, as these areas are unevenly stretched in the panorama images.
\section{Additional Ablation Study}
\label{sec:ablation}
\setcounter{table}{0}
\setcounter{figure}{0}

\begin{figure}[t]
  \centering
    \includegraphics[width=\linewidth]{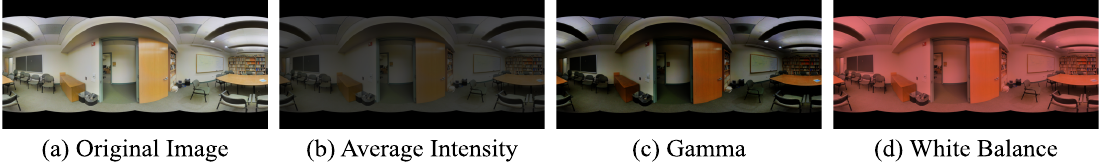}
    \caption{Synthetic color variations for evaluating illumination robustness.}
   \label{fig:color}
\end{figure}

\begin{table}[t]
    \centering
    \caption{Ablation study on color preprocessing evaluated in a subset of Stanford 2D-3D-S~\cite{stanford2d3d}. The images are modified by average intensity (Int.), gamma (Gam.), and white balance (W.B.).}
    \resizebox{\linewidth}{!}{
    \begin{tabular}{l|cccc|cccc|cccc}
        \toprule
        {} & \multicolumn{4}{c|}{$t$-error (m)} & \multicolumn{4}{c|}{$R$-error ($^\circ$)} & \multicolumn{4}{c}{Accuracy}\\
        Method & Orig. & Int. & Gam. & \begin{tabular}{@{}c@{}}W.B. \end{tabular} & Orig. & Int. & Gam. & \begin{tabular}{@{}c@{}}W.B. \end{tabular} & Orig. & Int. & Gam. & \begin{tabular}{@{}c@{}}W.B. \end{tabular} \\
        \midrule
        \begin{tabular}{@{}l@{}}CPO w/o \\ Preprocessing\end{tabular} & - & 3.85 & 3.48 & 3.40 & - & 153.92 & 136.96 & 129.05 & - & 0.00 & 0.00 & 0.03 \\ \midrule
        CPO & 0.01 & \textbf{0.01} & \textbf{0.01} & \textbf{0.01} & 0.19 & \textbf{0.21} & \textbf{0.25} & \textbf{0.25} & 0.94& \textbf{0.88} & \textbf{0.88} & \textbf{0.88}\\
        \bottomrule
    \end{tabular}
    }
    \label{table:illumination}
\end{table}

\paragraph{Color Preprocessing for Illumination Robustness} 
We report the impact of preprocessing the color values of the panorama and point cloud for robustness against illumination changes.
Recall that we match the color distributions of 2D and 3D via optimal transport, as mentioned in Section 3.1.
We apply synthetic color variations to the subset of images in Area 3 from Stanford 2D-3D-S~\cite{stanford2d3d}, as shown in Figure~\ref{fig:color}.
These images are originally used for obtaining results in Table 4 to make comparisons between CPO, PICCOLO, and GOSMA~\cite{gosma}.

We consider three synthetic color variations: average intensity, gamma, and white balance change.
For average intensity change we lower each pixel intensity by 33\%.
For gamma change, we set the image gamma to 3.
For white balance change, we apply the following transformation matrix to the raw RGB color values: $\begin{pmatrix}
1 & 0 & 0\\
0 & 0.5 & 0\\
0 & 0 & 0.5
\end{pmatrix}$.

Table~\ref{table:illumination} shows the results for illumination robustness.
CPO using color preprocessing shows robust performance amidst the three variations, whereas CPO without color distribution matching leads to poor performance in illumination changes.
While more sophisticated color modification methods~\cite{C5,style_transfer_1,style_transfer_2,style_transfer_3} may account for complex illumination shifts, we find that our simple matching scheme suffices for handling modest color variations in practical settings.

\section{Additional Qualitative Results}
\label{sec:qualitative}
\setcounter{table}{0}
\setcounter{figure}{0}

\begin{figure}[t]
  \centering
   \includegraphics[width=0.9\linewidth]{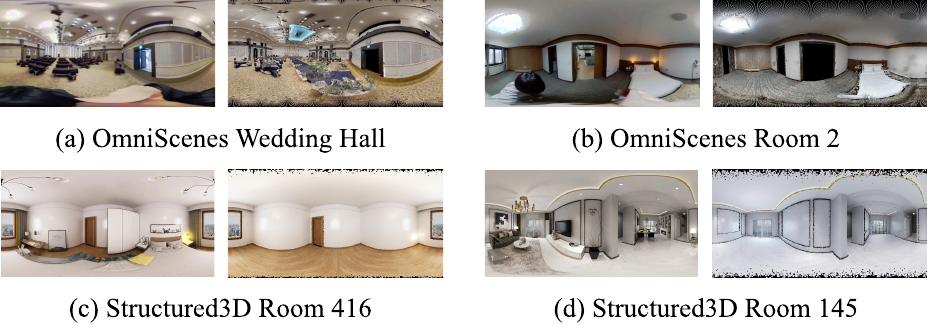}

\caption{Qualitative results of CPO on OmniScenes~\cite{piccolo} and Structured3D~\cite{Structured3D}. We display the input query image (left) and the projected point cloud under the estimated camera pose (right).}
\label{fig:qualitative}
\end{figure}

\begin{figure}[t]
  \centering
   \includegraphics[width=0.9\linewidth]{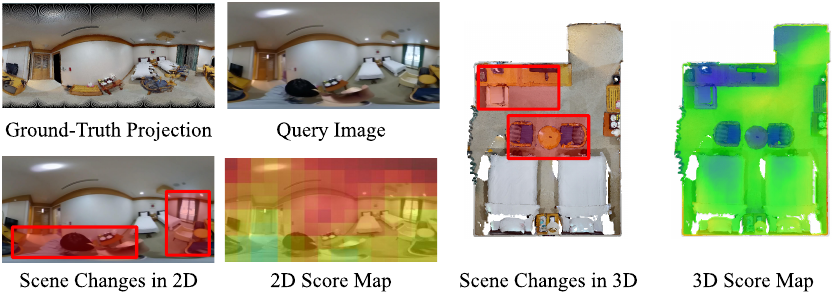}

   \caption{Visualization of 2D, 3D score maps. The 2D score map assigns lower scores to the capturer's hand and dislocated objects. Similarly, the 3D score map assigns lower scores to dislocated chairs and tables.}
   \label{fig:visscore}
\end{figure}

\paragraph{Localization in Scenes with Changes}
We further report additional qualitative results of CPO in OmniScenes~\cite{piccolo} and Structured3D~\cite{Structured3D}.
As shown in Figure~\ref{fig:qualitative}, CPO performs robust localization under various scenes in both datasets containing large amounts of scene change.

\paragraph{2D, 3D Score Maps}
We display additional 2D and 3D score maps generated for room 4 from OmniScenes~\cite{piccolo}.
As shown in Figure~\ref{fig:visscore}, the object arrangements have changed since the 3D scan.
Both 2D and 3D score maps assign smaller scores to dislocated objects and the 2D score map further attenuates capturer’s hand, which is not present in the 3D scan.
The score maps effectively place smaller weight on regions with scene changes, leading to robust localization in CPO as demonstrated in Section 4.

%
%

%
%
\bibliographystyle{splncs04}
\bibliography{egbib}
\end{document}